# Towards Medical Knowmetrics: Representing and Computing Medical Knowledge using Semantic Predications as the Knowledge Unit and the Uncertainty as the Knowledge Context

Xiaoying Li[1#], Suyuan Peng[2#], Jian Du[2*]

[1] Institute of Medical Information, Chinese Academy of Medical Sciences, Beijing, China

[2] National Institute of Health Data Science, Peking University, Beijing, China

**Abstract:** In China, Prof. Hongzhou Zhao and Zeyuan Liu are the pioneers of the concept "knowledge unit" and "knowmetrics" for measuring knowledge. However, the definition of "computable knowledge object" remains controversial so far in different fields. For example, it is defined as 1) quantitative scientific concept in natural science and engineering, 2) knowledge point in the field of education research, and 3) semantic predications, i.e., Subject-Predicate-Object (SPO) triples in biomedical fields. The Semantic MEDLINE Database (SemMedDB), a high-quality public repository of SPO triples extracted from medical literature, provides a basic data infrastructure for measuring medical knowledge. In general, the study of extracting SPO triples as computable knowledge unit from unstructured scientific text has been overwhelmingly focusing on scientific knowledge per se. Since the SPO triples would be possibly extracted from hypothetical, speculative statements or even conflicting and contradictory assertions, the knowledge status (i.e., the uncertainty), which serves as an integral and critical part of scientific knowledge has been largely overlooked. This article aims to put forward a framework for *Medical Knowmetrics* using the SPO triples as the knowledge unit and the uncertainty as the knowledge context. The lung cancer publications dataset is used to validate the proposed framework. The uncertainty of medical knowledge and how its status evolves over time indirectly reflect the strength of competing knowledge claims, and the probability of certainty for a given SPO triple. We try to discuss the new insights using the uncertainty-centric approaches to detect research fronts, and identify knowledge claims with high certainty level, in order to improve the efficacy of knowledge-driven decision support.

## 1. Introduction

Knowledge metrics, or knowmetrics, or epistometrics (a form of epistemology) is an interdisciplinary research fields on measuring knowledge instead of information (Hou et al., 2009) (Galyavieva, 2013). In general, metrics are defined as a combined whole of quantitative techniques associated with a specialized field of scientific knowledge, in order to obtain descriptive, evaluative or prospective results from the activities or phenomena analysis (Cavaller, 2008). In contrast to other metric disciplines such as bibliometrics, scientometrics, and informetrics, the concept of knowmetrics was initially introduced for the quantitative analysis on "element of knowledge" by Chinese scientometricians Prof. Hongzhou Zhao and Guo Hua Jiang (Zhao & Jiang, 1984), as well as Prof. Zeyuan Liu (Liu, 1999). Taking the entire system of human knowledge as its research subject, knowmetrics is an emerging subject that carries out a comprehensive study of knowledge capacity of the society and the social connection of knowledge through such methods as quantitative

---

analysis and computing technology. However, this definition only covers the general research paradigms based on traditional approach in branches of sciences and scientometrics. It involves little of the methodology of measuring knowledge units, which is the key to knowmetric research (Hou et al., 2009).

In fact, there is no consensus on the definition of knowmetrics. We think this concept has evolved and been promoted by three communities generally. The first is from the macro-level by the *science of science* community, who concentrates on knowmetrics and its application of the measurement of knowledge economies. The second is from the meco-level by the community of scientometrics and informatics, who focus on the science knowledge graph for mapping science domains and measuring knowledge structure (Borner, Chen, & Boyack, 2003; C. M. Chen, 2006). And the third is from the micro-level by the medical informatics community who combined the medical knowledge organization systems (such as MeSH, UMLS) and the computational techniques for new knowledge discovery in biomedical sciences (Bakal, Talari, Kakani, & Kavuluru, 2018; Keselman et al., 2010). Both the second and the third community's concerns are scientific publications mining. Specifically, the former focuses on the analysis of bibliographical data from scientific publications, while the latter emphasizes on 1) the "knowledge unit", in terms of Subject-Predicate-Object (SPO) triples extracted from the scientific text (Kilicoglu, Rosemblat, Fiszman, & Shin, 2020), or 2) the "computable knowledge object", expressed in code such as disease prediction models, learned from big data (Flynn, Friedman, Boisvert, Landis-Lewis, & Lagoze, 2018; Friedman & Flynn, 2019). The Semantic MEDLINE Database (SemMedDB), a high-quality public repository of SPO triples extracted from medical literature, provides a basic data infrastructure for measuring medical knowledge at the level of knowledge units (Kilicoglu, Shin, Fiszman, Rosemblat, & Rindflesch, 2012). In general, the study of extracting SPO triples as computable knowledge unit from unstructured scientific text has been overwhelmingly focusing on scientific knowledge per se. However, the status of scientific knowledge, i.e., the uncertainty, which serves as an integral and critical part of scientific knowledge, has been largely overlooked.

Uncertain scientific knowledge refers to the knowledge comes from hypothetical, speculative statements, or even conflicting and contradictory assertions, which are critical to understand the incremental and transformative development of scientific knowledge. The study and measurement of uncertainty of scientific knowledge and how uncertainty was expressed in scientific texts opens up a new area in the study in scientometrics and informatics (C. Chen & Song, 2017b). (Evans & Foster, 2011) introduced the concept of "metaknowledge", i.e., knowledge about knowledge. The growth of electronic publication and informatics archives makes it possible to harvest vast quantities of metaknowledge. The computational production and consumption of metaknowledge will allow researchers and policymakers to leverage more scientific knowledge - explicit, implicit, and contextual, in their efforts to advance science, such as recalibrate scientific certainty in particular propositions. Inspired by (Kuhn & Hacking, 2012), the evolutionary process of science can also be understood from the perspective of uncertainty of knowledge, which is evolving over time. Essentially, incremental science can be understood for a given unsolved scientific issue progressing from a wholly unknown state, to hypotheses or speculations, and then verifying the uncertainty surrounding the hypothesis, until reaching a reasonable conclusion which has a considerable proportion of evidence. Transformative science involves raising conflicting and contradictory interpretations with previous knowledge, leading to scientific disputes, and steadily promoting revolution. One extreme of uncertainty is "entirely unknown"; the other is "generally accepted as fact". There are mainly two forms of expression in between: one is hypothesis and speculation (i.e., hedging in language), the other is contradictory and controversy, corresponding to incremental and transformative development of science, respectively.

Uncertain knowledge is particularly common in medical sciences. Recently, two independent studies investigated the frequency of textual uncertainty cues, such as "may", "might", "could", as well as "controversial", "contradictory", and "conflicting", in all sentences or only citing sentences in Elsevier's full text database. They have found that after social sciences, medical sciences rank second according to the frequency of uncertain text, and medical sciences have the most "disagreement", i.e., controversial scientific claims (C. Chen, Song, & Heo, 2018; Murray et al., 2019). It is not rare in medical practices to encounter "medical reversals", in which prior studies that claimed some therapeutic benefit were contradicted by subsequent research (Prasad et al., 2013). Through an analysis of more than 3000 randomized controlled trials (RCTs) published in three leading medical journals (*the Journal of the American Medical Association*, *the Lancet*, and *the New England Journal of Medicine*), (Herrera-Perez et al., 2019) have identified 396 medical reversals. The reality in medical practice is that doctors continually make decisions on the basis of imperfect data and limited knowledge, which may lead to diagnostic uncertainty, coupled with the uncertainty that arising from unpredictable patient responses to treatment and from health care outcomes that are far from binary. (Simpkin & Schwartzstein, 2016) believe that a shift toward the acknowledgment and acceptance of uncertainty of medical knowledge is essential - for physicians, for patients, and for health care system as a whole.

Scientific publications can be seen as records of knowledge claims on a research question, supported by empirical evidence. By examining the linguistic characteristics exposed in scientific texts, it is possible to quantitatively measure the uncertainty of scientific knowledge. Removing redundant part of scientific text and extracting structured knowledge unit is the key to realize intelligent knowledge mining and promote knowledge translation from research to action, but this process often ignores the representation of knowledge status, i.e., the uncertainty. This article aims to propose a framework for "*Medical Knowmetrics*" using the SPO triples as the knowledge unit and the uncertainty as the knowledge context. The lung cancer publications dataset is used to validate the framework.

## 2. Related Work

We summarized recent advances from two aspects. The first is to identify uncertain cue words in biomedical scientific text; the second refers to extracting machine-understandable knowledge from unstructured biomedical text.

### 2.1 Identifying uncertain cue words in biomedical scientific text

The research on identifying uncertainty expressions in scientific literature started in the late 1990s (Hyland, 1996). It was first used in the scientific writing field for analyzing 63 hedging cue words, which indicate that the author's scientific judgments are subjectively cautious and cover about 11% of the sentences in PubMed abstracts (Light et al., 2004). During 2010s, it was continually investigated by the community of computational linguistics (Zerva, 2019; Thompson et al., 2011; Szarvas et al., 2012), which treated hedging as a linguistic phenomenon and used computer science for automatic recognition. In the past three years, the community of informatics has begun to converge on the measurement of uncertainty in science and pay more attention to the distribution of uncertainty cue words in full-text (Chen and Song, 2017b). They also extend the scope of textual uncertainty from only hedging cues by computational linguists to a broad-coverage of uncertainty such as controversial, inconsistence and contradiction contained in scientific text (Chen et al., 2018).

A few studies have been carried out using the full-text of scientific publications. For instance, (Mercer, Di Marco, & Kroon, 2004) took the full-text of the 985 BioMed Central papers as corpus and found that the hedging words described by (Hyland, 1996) are more likely to appear in citing sentences. Recently, using citing sentences and the cited publications in PubMed Central full-text database, (Small, 2018) proposed a measure called hedging rate of a given publication, which is defined as the proportion of citing sentences with the three most prominent hedging terms "may", "could" and "might" in all citing sentences. He found that method papers have a lower hedging rate than non-method papers, which means method paper in general has higher certainty than non-method paper. In general, rates of hedging are found to be higher for papers with fewer citances, suggesting that the certainty of scientific results is directly related to citation frequency, and early citing sentences will have a higher hedging rate than later citing sentences (Small, Boyack, & Klavans, 2019). By comparing word usage of citing sentences for low-hedged and highly hedged papers, it was found that low-hedged, or high certainty, papers were associated with action verbs denoting the application of methods and acquisition of data, and words specifically denoting quantitative methods (e.g., "using", "performed", …). High-hedged, or low certainty, papers, on the other hand, were associated with words denoting interpretation and justification of ideas (e.g., "suggest", "evidence", …) (Small, 2019). Most recently, (Small, 2020) suggests a new direction for quantitative science studies: it will be important for quantitative science studies to address confirmation in science and the role of evidence in that process by approaching confirmation as bibliometric problem. With citation contexts we can assess the collective opinions of the scientific community, using hedging to measure uncertainty, and epistemic words to locate causal assertions tied to evidence. One place to begin is to identify causal networks embedded in our maps of science, tracking them over time to assess the impact of evidence on the certainty or uncertainty of theoretical assumptions.

Disagreement or dispute is one type of uncertainty of scientific knowledge. Dispute in science is central to the production of new knowledge. (Murray et al., 2019) recently developed a methodology for investigating disagreement in science based on citing sentences from the Elsevier ScienceDirect database. They focus on citation sentences containing the disagreement signal phrases "conflict" or "contradict", which occur alongside filter phrases "studies" or "results". We noticed that the authors employed combinations of cue words to identify "disagreement" statements. Disagreement filter phrases must appear within a four-word window of the signal. (Atanassova, Rey, & Bertin, 2018) also argued that the task of identifying uncertain sentences cannot be accomplished by recognizing cue words only. And a terms-combination strategy is also recommended for the detection of speculative statements in scientific text (Malhotra, Younesi, Gurulingappa, & Hofmann-Apitius, 2013). Supplementing additional words as filters will enhance the precision of uncertain sentence recognition.

## 2.2 Extracting machine-executable knowledge from unstructured biomedical text

Firstly, the advancement in semantic knowledge representation provides insights for the research to be carried out in this paper. SemRep is a rule-based natural language processing system, based on 1) medical concepts, 2) concept types (e.g., drugs, diseases), and 3) semantic relationships between concepts (e.g., drugs-TREAT-diseases) in Unified Medical Language System (UMLS). Using SemRep, one can extract SPO triples from biomedical text. In 2012, the project team developed SemMedDB to provide a basic data infrastructure for measuring medical knowledge, which stored the semantic predications and the corresponding sentences extracted from the title and abstract of PubMed publications based on SemRep (Kilicoglu et al., 2012). Although SemRep and SemMedDB accomplish the extraction and storage of large-scale structured knowledge units, they lose most of the meta-knowledge, especially the uncertainty status of scientific assertions. This enables the development of computable medical knowledge units in this work.

The second is the "nano-publication" model (Groth, Gibson, & Velterop, 2010). The meaning of "nano" here refers to the smallest knowledge unit which is machine-readable. It includes three parts: (a) assertion expressed by subject-predicate-object triples, (b) source information, and (c) publication information, which is the metadata about nano-publication itself, including the creator, creation date and the version of the nano-publication. The third is a micropublication model proposed by (Clark, Ciccarese, & Goble, 2014). The minimal form of a micropublication is a statement with its attribution. The maximal form is a statement with its complete supporting argument, consisting of all relevant evidence, interpretations, discussion and challenges brought forward in support of or opposition to it.

However, each of the above three models have advantages and disadvantages. First, SemRep and SemMedDB accomplish the extraction and storage of large-scale structured knowledge units, but lose most of the meta-knowledge, especially the uncertainty status of scientific assertions. For example, scholars tend to modify their conclusions to express them as objectively as possible. Assertions may be speculations and not facts, and they even may come from conflicting, contradictory, and controversial statements. It's the same as nano-publication, which enables computable knowledge units that can be traced by assigning unique IDs. However, it does not indicate whether the scientific assertion is a conclusive claim from the current new experiment, or only cite a previous scientific assertion, or simply a hypothetical statement. Second, while micropublications enable us to formalize the arguments and evidence in scientific publications, but the arguments, or the core assertions, are statements in natural language sentence and are not structured thus not computable.

## 3. Methodology and Framework

### 3.1 A representation model for computable medical knowledge objects

The framework is built on a key aspect of scientific papers: claims. Claims are natural language sentences in a scientific paper that expresses a relationship between two entities. In particular, how one of them affects, manipulates, or causes the other entity. Firstly, we introduce the concept of basic knowledge unit (Figure 1), that is one SPO triple derived from the aggregation of multiple source sentences. And we propose a representation model for computable medical knowledge (Figure 2), which consists of four elements: 1) the unique ID, 2) basic knowledge unit, 3) supporting sentences as the knowledge sources, and 4) uncertainty as the knowledge status.

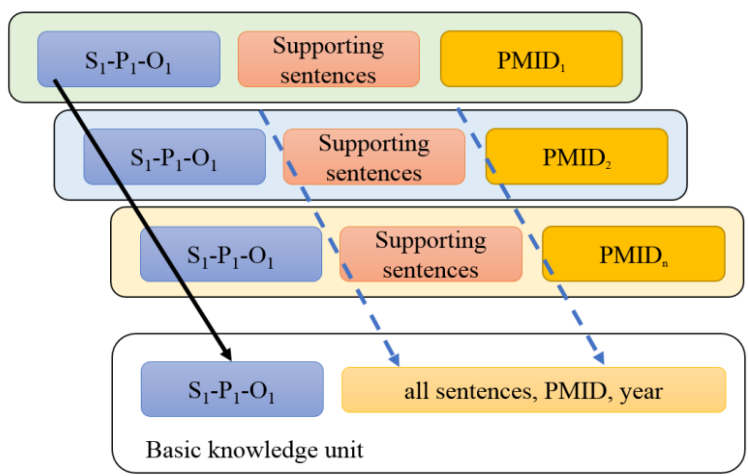

Figure 1. The concept of basic knowledge unit

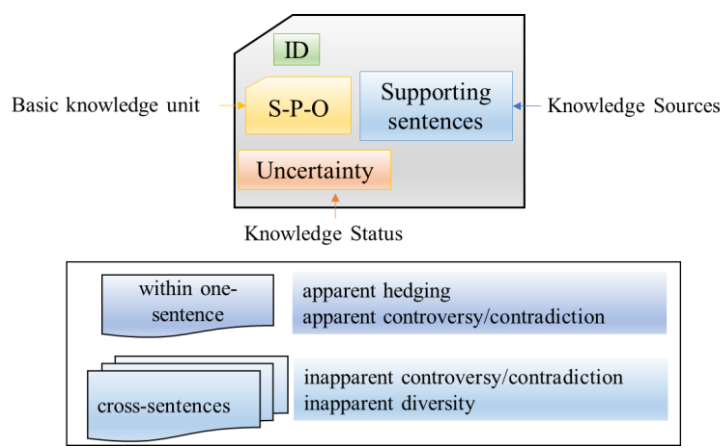

Figure 2. A representation model for computable medical knowledge objects

We classified the uncertainty of medical knowledge into three levels, 'hedging', 'diversity' and 'controversy/contradiction'. Hedging is a particularly relevant concept in understanding how scientists characterize the tentative and context-dependent nature of scientific claims. Commonly used hedging words include *"may"*, *"could"*, and *"might"*. Diversity refers to such claims that connect the same entities but with different associations that not necessarily contradict each other. The idea here is to capture the presence of scientific claims that may deserve further investigation because they have different semantics although they share almost the same context. The difference between contradiction and controversy is that contradiction is the act of contradicting while controversy is a debate, and discussion of opposing opinions and conflicting results. The case of controversy/contradiction refers to that there are at least two claims with the same pairs of entities that have opposite semantic orientation. The textual uncertainty can be detected 1) within one-sentence to identify apparent hedging using cue words such as *may, could,* and *might*, as well as apparent controversy/contradiction using cue words such as *conflicting, controversial,* and *contradictory*, and 2) from cross-sentences to identify inapparent diversity, and inapparent controversy/contradiction using the semantic orientation of knowledge claims in the documents.

### 3.2 A framework for extracting uncertain knowledge

Our classification here focuses on two core uncertain aspects: 1) controversy/contradiction, and 2) diversity. Controversy arises when two or more claims semantically contradict each other; diversity means the presence of different semantics of the claims that do not contradict each other but provide different insights. We use SemRep to extract SPO triples in abstracts of PubMed publications, and store the SPO triples together with their corresponding sentences in our local databases, which formed the basis of our study.

*3.2.1 Using SemRep to extract SPO triples from sentences in abstracts of PubMed publications*

SemRep is a well-known knowledge-based semantic relation interpreter developed by the National Library of Medicine (NLM), with semantic relation structured in the form of SPO triple. Briefly, SemRep identifies semantic triples by interpreting the scientific text, which exploits syntactic analysis in natural language processing and structured domain knowledge from UMLS with three knowledge sources. Currently, SemRep facilities the semantic relation extraction from PubMed database by setting MEDLINE format as default of input text, which ensures it a powerful tool for knowledge mining. SemRep has not been formally evaluated, due to the lack of a gold standard corpus; however, many task-based evaluations have been reported that SemRep generates some errors with its precision scores varied between 75-96% (Rindflesch and Fiszman, 2003; Kilicoglu et al., 2012).

*3.2.2 Identifying knowledge with controversy/contradiction*

We use two approaches to discovering controversial or contradictory knowledge. The first is directly searching such sentences which contain three prominent cue terms "controversial", "contradictory", and "conflicting" inspired by (Murray et al., 2019). The second approach is using the semantic orientation on the rule that, 1) two or more SPO triples are extracted from scientific claims, and 2) they have same subject (e.g., drug) and object (e.g., disease) but opposite predicates. In order to define opposite predicates, we classify the predicates into 2 groups: Excitatory and Inhibitory (Table 1). Here, group E lists the excitatory relations, while group I records the inhibitory relations. Generally, a knowledge claim $C_1$-$R_1$-$C_2$ is considered to be contradictory with another knowledge claim $C_1$-$R_2$-$C_2$ under the following cases:

- $R_1$ is an excitatory relation label (for example CAUSES) and $R_2$ is an inhibitory one (for instance PREVENTS);
- $R_1$ is an inhibitory relation label (for example TREATS) and $R_2$ is an excitatory one (for instance NEG_PREVENTS).

Table 1. Excitatory and Inhibitory semantic predicates

| Excitatory relations (Group E) | AUGMENTS, CAUSES, COMPLICATES, PREDISPOSES, PRODUCES, STIMULATES, NEG_DISRUPTS, NEG_INHIBITS, NEG_PREVENTS, |
|---|---|

|  | NEG_TREATS |
|---|---|
| **Inhibitory relations (Group I)** | DISRUPTS, INHIBITS, PREVENTS, TREATS, NEG_AUGMENTS, NEG_CAUSES, NEG_COMPLICATES, NEG_PREDISPOSES, NEG_PRODUCES, NEG_STIMULATES |

Table 2 shows a pair of knowledge with contradiction, where the first semantic predication contradicts the last one due to contradictory predicates "PREDISPOSES" and "NEG_PREDISPOSES".

Table 2. A pair of contradictory knowledge

| Scientific claim | Subject | Predicate | Object |
|---|---|---|---|
| 21862624.ab.11 The directions of association for 15q25 variants with *cotinine* were in *accordance* with that expected of *lung cancer risk*. | Cotinine | PREDISPOSES | Malignant neoplasm of lung |
| 15681570.ab.12 *Cotinine* concentration was clearly associated with self reported exposure (3.30, 2.07 to 5.23, for detectable/non-detectable cotinine), but it was *not associated* with the *risk* of respiratory diseases or *lung cancer*. |  | NEG_PREDISPOSES |  |

Taking advantage of semantic predications from SemRep, the discovery of contradictory knowledge between drug and disease could be expressed by the formula below.

| **Algorithm 1:** Contradictory knowledge discovery |
|---|
| **Input:** Semantic predications $C_1$-$P_1$-$C_2$ and $C_1$-$P$-$C_2$ from scientific claims |
| **Output:** Contradictory knowledge ($C_1$, $P_1$, $C_2$) and ($C_1$, $P_2$, $C_2$), where $P_1$ and $P_2$ meet one of the following conditions:<br>1: $P_1 \in E$ & $P_2 \in I$;<br>2: $P_1 \in I$ & $P_2 \in E$. |

### 3.2.3 Identifying knowledge with diversity

For two or more SPO triples interpreted by SemRep from scientific claims, if 1) they have identical subject (e.g., drug) and object (e.g., disease), but various predicates; 2) all of these predicates belong to either group E or group I and do not constitute a contradiction, then we will consider the knowledge concerning a given pair with diversity. Table 3 gives an example of drug-disease knowledge with diversity.

Table 3. An example of diverse knowledge

| Scientific claim | Subject | Predicates | Object |
|---|---|---|---|
| 22684632.ab.8 *Selenium* showed beneficial effects on gastrointestinal cancer and *reduced the risk of lung cancer* in populations with lower selenium status. | Selenium | PREVENTS | Malignant neoplasm of lung |
| 22073154.ab.3 Two independent reviewers searched six databases from inception to March 2009 for evidence pertaining to the safety and *efficacy* of *selenium for lung cancers*. |  | TREATS |  |

The new algorithm for diverse knowledge discovery is formulated as follows.

**Algorithm 2:** Diverse knowledge discovery

**Input:** Semantic predications $C_1$-$P_1$-$C_2$ and $C_1$-$P_2$-$C_2$ from scientific claims

**Output:** Diverse knowledge ($C_1$, $P_1$, $C_2$) and ($C_1$, $P_2$, $C_2$) which satisfies both of the following two conditions:

1: $\nexists P_2$ where ($C_1$, $P_2$, $C_2$) and ($C_1$, $P_1$, $C_2$) contradict each other;

2: ($P_1 \in$ E & $P_2 \in$ E), or ($P_1 \in$ I & $P_2 \in$ I).

## 4. Results

Based on the proposed framework which uses semantic predications as the knowledge unit and the uncertainty as the knowledge context, we try to extract and characterize the uncertain knowledge on the field of drug treatment for lung cancer (Figure 4).

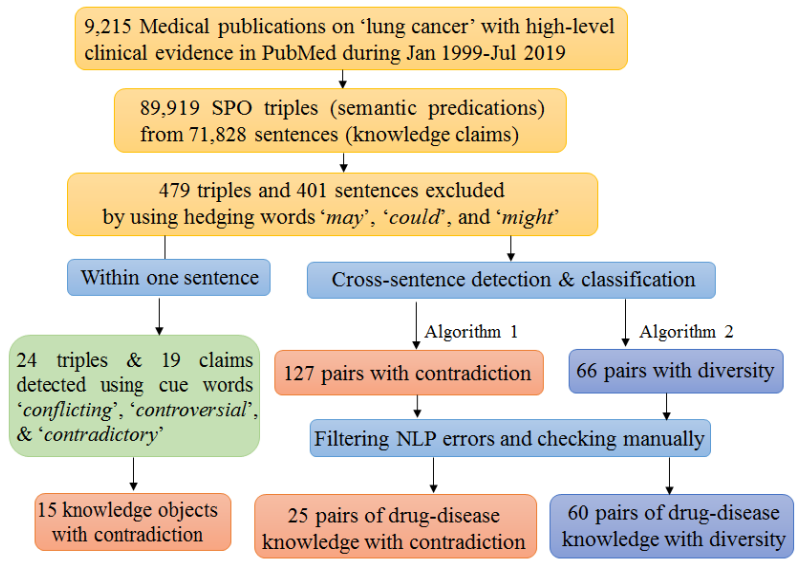

Figure 4. The experimental flowchart of our study

### 4.1 Selecting medical publications with high-level clinical evidence

Different types of studies providing different levels of evidence. In a real-world application, it may be more preferable to assign lower weight to a contradiction due to a case study finding compared to a contradiction between the findings of two Randomized Controlled Trials or systematic reviews (Rosemblat, Fiszman, Shin, & Kilicoglu, 2019). To extract the computable knowledge units from supporting sentences and discover the contradictory knowledge with higher confidence and precision, the PubMed publications with high-level clinical evidence were queried and selected using both the information of Publication Type (PT) and Medical Subject Headings (MeSH) according to (Haynes, 2006):

Publication Types (PT):

"Guideline" OR "Practice Guideline" OR "Meta-Analysis" OR "Multicenter Study" OR "Randomized Controlled Trial" OR "Clinical Trial" OR "Clinical Trial, Phase I" OR "Clinical Trial, Phase II" OR "Clinical Trial, Phase III" OR "Clinical Trial, Phase IV" OR "Pragmatic Clinical Trial" OR "Comparative Study" OR "Controlled Clinical Trial"

Medical Subject Headings (MeSH):

"Meta-Analysis as Topic" OR "Randomized Controlled Trials as Topic" OR "Systematic

*Reviews as Topic" OR "Clinical Trials as Topic" OR "Clinical Trials, Phase I as Topic" OR "Clinical Trials, Phase II as Topic" OR "Clinical Trials, Phase III as Topic" OR "Clinical Trials, Phase IV as Topic"*

We collected 9,215 medical abstracts on "lung cancer" published during Jan 1999-Jul 2019 (Figure 4). After locally preservation, we run the batch mode of SemRep for semantic predications (SPO triples) extraction, and 89,919 SPO triples were extracted from 71,828 sentences (knowledge claims). To restrict generous biomedical concepts within drug and disease, we make proper use of UMLS knowledge by selecting the semantic predications with the subject and object holding semantic types "Chemicals and Drugs" and "Disease or Syndrome". Besides, the hierarchy of "Pharmacologic Actions" from MeSH thesaurus was adopted to automatically filter out the generic concepts of drugs, such as "Antineoplastic Agents", "Anti-Inflammatory Agents".

**4.2 Filtering SPO triples supported by hedging sentences**

In order to discover the contradictory knowledge with higher confidence and precision, we automatically filter out the semantic predications interpreted from uncertain claims by using hedges. Generally, if the claims in scientific texts contain hedging terms, then we can infer the authors consider part of their work as uncertain in some respect. Consequently, the semantic predications interpreted from these uncertain claims will be regarded with uncertainty. For our purpose, we do not need to identify all the hedges in scientific claims, only the three most prominent ones, namely "may", "could", and "might" were selected according to (Small et al., 2019). Through this process, 479 triples were filtered out from 401 sentences containing these three hedges.

**4.3 Extracting uncertain knowledge within one sentence**

In general, 24 triples and 19 claims were detected by using cue words 'conflicting', 'controversial', and 'contradictory' within one sentence. Two of them were not in the scope of the present study, because the subjects were not a drug (nor drug candidate compound) (Table 4). Here, the chemotherapy agents had the highest number of apparent contradictions among all agents (11; 64.7%). During the 1970s to 1980s, the majority of oncological trials concentrated on evaluating response rates, disease‐free intervals, and overall survival as endpoints. In recent decades, the perceived role of chemotherapy in the treatment of advanced cancers has been changed. Palliative care is promoted as an approach to improve quality. Palliative chemotherapy has been the backbone of therapy in advanced-stage disease and has evolved over time. Moreover, chemotherapy side effects elimination while keeping the original therapeutic advantages has been of great interest over the last decades. The need for better evidence for the most effective type of chemotherapy regimen can lead to medical reversals and suggest that reality checks should be encouraged for established practices to avoid subjectivity and medical inertia.

Table 4. Knowledge detected by using cue words 'conflicting', 'controversial', and 'contradictory' within one sentence

| Date of Publication | Knowledge source (PMID-claims) | Basic knowledge unit (S-P-O triple) | Knowledge status (uncertainty) |
|---|---|---|---|
| 1999 Oct | 10528024.ab.1 BACKGROUND: Higher blood levels of alpha-tocopherol, the predominant form of vitamin E, have been associated in some studies with a reduced risk of lung cancer, but | Vitamin E-DISRUPTS-Malignant neoplasm of lung | *conflicting* |

| Date of Publication | Knowledge source (PMID-claims) | Basic knowledge unit (S-P-O triple) | Knowledge status (uncertainty) |
|---|---|---|---|
| | other studies have yielded *conflicting* results. | | |
| 2008 Jan | 18187393.ab.1 BACKGROUND: Prior studies indicate that use of aspirin or other nonsteroidal anti-inflammatory drugs (NSAID) is associated with a decreased risk of non-small cell lung cancer (NSCLC); however, results have been *contradictory* in part because of variation in study design. | Aspirin-PREVENTS-Non-Small Cell Lung Carcinoma | *contradictory* |
| 2009 Sep | 19144444.ab.1 OBJECTIVE: Whether platinum plus gemcitabine or vinorelbine are equally effective in the treatment of advanced non-small-cell lung cancer (NSCLC) is still *controversial*, a meta-analysis was performed to compare the gemcitabine plus platinum | Gemcitabine, vinorelbine-TREATS- Non-Small Cell Lung Carcinoma | *controversial* |
| 2011 Nov | 22073154.ab.1 BACKGROUND: Selenium is a natural health product widely used in the treatment and prevention of lung cancers, but large chemoprevention trials have yielded *conflicting* results. | Selenium-PREVENTS-Malignant neoplasm of lung | *conflicting* |
| 2014 Feb | 24372368.ab.1 PURPOSE: Our knowledge on bronchoalveolar lavage (BAL) of methotrexate-induced pneumonitis (MTX-P) is fragmentary and based on data that are sometimes apparently *conflicting*. | Methotrexate-CAUSES-Pneumonia | *conflicting* |
| 2014 Oct* | 24989113.ab.2 Impacts on the catalytic activity of the CYP1B1 enzyme, as well as an association of the Leu432Val polymorphism with the risk of lung cancer, have been described; however, the results remain *controversial*. | Cytochrome-P-450-CYP1B1 PREDISPOSES Malignant neoplasm of lung | *controversial* |
| 2014 Aug | 25016505.ab.1 PURPOSE: Evidence on the benefits of combining celecoxib, a cyclooxygenase-2 inhibitor, in treating advanced cancer is still *controversial*. | Celecoxib-TREATS-Advanced cancer | *controversial* |
| 2014 Jul | 25029199.ab.1 BACKGROUND: Since efficacy and safety of epidermal growth factor receptor tyrosine kinase inhibitors (EGFR-TKIs) versus chemotherapy in the treatment of patients with pretreated advanced non-small cell lung cancer (NSCLC) remain *controversial*, we performed a meta-analysis to compare them. | Epidermal growth factor receptor inhibitor (product) –TREATS-Non-Small Cell Lung Carcinoma | *controversial* |
| 2014 Mar | 25818741.ab.1 BACKGROUND: The extent of the benefit of gefitinib in the treatment of advanced nonsmall-cell lung cancer (NSCLC) is till *controversial*, when compared with docetaxel. | Gefitinib-TREATS-Non-Small Cell Lung Carcinoma | *controversial* |
| 2015 Nov | 26548932.ab.1 BACKGROUND: The extent of the benefit of erlotinib in the treatment of advanced nonsmall-cell lung cancer (NSCLC) is still *controversial* when compared with docetaxel. | Erlotinib-TREATS-Non-Small Cell Lung Carcinoma | *controversial* |
| 2016 Aug | 27305276.ab.1 BACKGROUND/AIM: The utility of pulmonary function testing (PFT) to detect bleomycin-induced pneumonitis is *controversial*. | Bleomycin-CAUSES-Pneumonia | *controversial* |
| 2017 Apr | 28150074.ab.1 Background The efficacy and safety of bevacizumab in elderly patients with non-small cell lung cancer remain *controversial*. | Bevacizumab-TREATS-Non-Small Cell Lung Carcinoma | *controversial* |

| Date of Publication | Knowledge source (PMID-claims) | Basic knowledge unit (S-P-O triple) | Knowledge status (uncertainty) |
|---|---|---|---|
| 2017 Jun* | 28502040.ab.3 Whether UGT1A1*6 and UGT1A1*28 are associated with IRI-induced neutropenia, diarrhea and IRI-based chemotherapy tumor response (TR) in Asians with lung cancer remains *controversial*. | Irinotecan-CAUSES-Leukopenia | *controversial* |
| 2017 Jun | 28643733.ab.1 BACKGROUND: The use of cisplatin (Cis) versus carboplatin (Carb) in the treatment of advanced nonsmall cell lung cancer (NSCLC) is *controversial*. | Cisplatin-TREATS-Non-Small Cell Lung Carcinoma | *controversial* |
| 2017 Jun | 28643733.ab.1 BACKGROUND: The use of cisplatin (Cis) versus carboplatin (Carb) in the treatment of advanced nonsmall cell lung cancer (NSCLC) is *controversial*. | Carboplatin-TREATS-Non-Small Cell Lung Carcinoma | *controversial* |

* Not in the scope of the present study.

## 4.4 Extracting uncertain knowledge from cross-sentences

We then implemented the presented algorithm 1 & 2 to automatically discover biomedical knowledge with contradiction and diversity. Finally, 127 candidate pairs of drug-disease knowledge with controversy and 66 candidate pairs with diversity were discovered for further analysis and evaluation.

*4.4.1 Filtering NLP errors and checking manually*

The manual biocuration aims to identify the accuracy of candidate knowledge based on the automatic algorithms, in terms of manually validation of the semantic predications and their supporting claims. In detail, two tasks involved in this process:

- Check whether the drug and disease were well recognized by MetaMap, which was already embedded in SemRep to map biomedical text into UMLS metathesaurus (e.g., Named Entity Recognition error (NER error), refer to the first row in Table 5);
- Verify whether SemRep's predicates properly indicated the semantic interactions between given drug and disease involved in supporting claims (e.g., Semantic Relation Extraction error (SRE error), see the last row in Table 5).

Table 5. Examples of inaccurate semantic predications from SemRep

| Scientific claim | Subject | Predicate | Object | Category |
|---|---|---|---|---|
| 11479851.ab.1 BACKGROUND/PURPOSE: *Antibody* to vascular endothelial growth factor (anti-VEGF) suppresses tumor growth and metastasis in experimental Wilms tumor. | *Vascular Endothelial Growth Factors* | DISRUPTS | tumor growth | NER error |
| 16152626.ab.11 In conclusion, these data *do not support* the hypothesis that intakes of vitamins A, C and E and folate reduce lung cancer risk. | Folate | *PREVENTS* | Malignant neoplasm of lung | SRE error |

To ensure consistence of biocuration in the annotation process, two authors (JD and XL) firstly established the criteria above, then one author (XL) annotated the candidate knowledge with contradiction and diversity, finally the other two authors (JD and PS) reviewed all the annotation results and made ultimate decision.

*4.4.2 Identified drug-disease knowledge with controversial/contradiction*

We found 25 groups of contradicted knowledge claims. Four of them were not a drug (nor drug candidate compound), which were beyond the scope of the present study. Our analysis of the contextual characteristics of 21 inapparent contradictions led to a categorization into seven main classes (Table 6 and 7). And the frequency of uncertain sentences across the whole dataset was counted and adopted as a measure to quantify the uncertainty of a specific SPO triple (see the third column of Table 6).

Table 6. The 25 pairs of drug-disease knowledge with contradiction

|  | Subject | Predicate (# claims) | Object | Category |
|---|---|---|---|---|
| Topic - 1 | Benzo(A)Pyrene | CAUSES (3) <br> NEG_CAUSES (1) <br> AUGMENTS (1) | Carcinogenesis | Category - 1 |
| Topic - 2 | Beta Carotene | NEG_PREVENTS (3) <br> PREVENTS (1) <br> PREDISPOSES (3) <br> AUGMENTS (2) | Malignant neoplasm of lung | Category - 2 |
| Topic - 3 | Carboplatin | PREVENTS (1) <br> TREATS (1) <br> CAUSES (1) | Thrombocytopenia | Category - 1, 7 |
| Topic - 4 | Cetuximab | TREATS (9) <br> NEG_TREATS (1) | Non-Small Cell Lung Carcinoma | Category - 3 |
| Topic - 5 | Cisplatin | TREATS (1) <br> NEG_TREATS (1) | Non-Squamous Non-Small Cell Lung Carcinoma | Category - 3 |
| Topic - 6 | Cisplatin | TREATS (5) <br> NEG_TREATS (1) | Small cell carcinoma of lung | Category - 4 |
| Topic - 7* | Cotinine | PREDISPOSES (1) <br> NEG_PREDISPOSES (1) | Malignant neoplasm of lung | / |
| Topic - 8 | Dacomitinib | TREATS (1) <br> NEG_TREATS (1) | Non-Small Cell Lung Carcinoma | Category - 1 |
| Topic - 9 | Docetaxel | TREATS (1) <br> CAUSES (1) | Febrile Neutropenia | Category - 1 |
| Topic - 10* | Epidermal Growth Factor Receptor | DISRUPTS (1) <br> CAUSES (1) | tumor growth | / |
| Topic - 11 | Erlotinib | NEG_AUGMENTS (1) <br> PREDISPOSES (1) | Lung Diseases, Interstitial | Category - 5 |
| Topic - 12 | Etoposide | TREATS (4) <br> NEG_TREATS (1) | Small cell carcinoma of lung | Category - 4 |
| Topic - 13 | Irinotecan | TREATS (3) <br> NEG_TREATS (1) | Small cell carcinoma of lung | Category - 4 |
| Topic - 14 | Isotretinoin | TREATS (1) <br> NEG_PREVENTS (2) | Malignant neoplasm of lung | Category - 6 |
| Topic - 15* | Matrix Metalloproteinase 2 | PREVENTS (1) <br> PREDISPOSES (1) | Malignant neoplasm of lung | / |
| Topic - 16 | Paclitaxel | TREATS (2) <br> NEG_TREATS (1) | extensive stage small cell lung cancer | Category - 1 |
| Topic - 17 | Particle | TREATS (2) <br> NEG_TREATS (1) | stage I non-small cell lung cancer | Category - 4 |
| Topic - 18 | Pemetrexed | TREATS (4) <br> NEG_TREATS (1) | Non-Small Cell Lung Carcinoma | Category - 1 |
| Topic - 19* | Rasagiline | TREATS (2) <br> CAUSES (1) | Malignant neoplasm of lung | / |
| Topic - 20 | Selenium | PREVENTS (1) <br> NEG_PREVENTS (1) | Malignant neoplasm of prostate | Category - 2 |

| | Subject | Predicate (# claims) | Object | Category |
|---|---|---|---|---|
| Topic - 21 | Tamoxifen | TREATS (1)<br>PREDISPOSES (1) | Malignant Neoplasms | Category - 1 |
| Topic - 22 | Vitamin E | PREVENTS (1)<br>NEG_PREVENTS (1) | Malignant neoplasm of prostate | Category - 1 |
| Topic - 23 | Anthracyclines | TREATS (3)<br>NEG_TREATS (1) | Carcinoma breast stage IV | Category - 1 |
| Topic - 24 | Capecitabine | TREATS (2)<br>NEG_TREATS (1) | Carcinoma breast stage IV | Category - 1 |
| Topic - 25 | Taxane | TREATS (2)<br>NEG_TREATS (1) | Carcinoma breast stage IV | Category - 1 |

\*, Not in the scope of the present study.

Table 7. Categories of contextual characteristics for explaining these contradictions

| Category | | Topic No. |
|---|---|---|
| Category - 1 | Heterogeneity in study design | 1, 3, 8, 9, 16, 18, 21, 22, 23, 24, 25 |
| Category - 2 | The contradiction of observational studies and RCTs | 2, 20 |
| Category - 3 | Research settings or Real word Settings | 4, 5 |
| Category - 4 | Cost-effectiveness | 6, 12, 13, 17 |
| Category - 5 | Short-term outcome | 11 |
| Category - 6 | Publication bias, citation bias and time-lag bias | 14 |
| Category - 7 | Inaccuracy of SemRep | 3 |

*(1) Heterogeneity in study design.* The contradictions between studies pertained to the complexity of methodological factors, such as different participants, interventions, durations, or sample sizes. Eleven groups of the identified contradictions were due to the heterogeneity in study design. The following examples have shown that the different dosage of agents' intervention affects the results. In particular, the heterogeneity associated with methodological diversity would indicate the results of studies have been biased estimated. Empirical evidence suggests that some aspects of the design can affect the result of clinical trials. Since medicine research is partly a statistically driven science, a certain amount of reversal of standards of care is inevitable. For example,

Topic – 16 [Paclitaxel – TREATS/NEG_TREATS - extensive-stage small cell lung cancer]

*The purpose of this study was to evaluate the **therapeutic effectiveness** of **paclitaxel** in previously untreated patients with **extensive-stage small-cell lung cancer (SCLC)**.* [PMID: 10521070]

*<u>Paclitaxel</u> in this dose and schedule should **not** be used as front-line **therapy** for patients with **ES-SCLC**.* [PMID: 18303437]

*(2) The contradiction of observational studies and RCTs.* Observational studies can test several hypotheses at a low cost in a short period of time in an epidemiological cohort representing the general population. While the weakness of observational studies is confounding bias, both the exposures and outcome events can be affected by confounders. Beta-carotene was initially supported by many epidemiological studies and laboratory investigations as potent chemoprevention against cancer. Nevertheless, RCTs involving supplementation of pharmacologic doses revealed the absence of beneficial effects and the potential for harm with beta-carotene use. Failing to account for potential confounders may lead the association of exposures and outcomes were over-/under-estimated, or may draw the opposite conclusion.

*(3) Research settings or real word settings.* Many medical reversals involve conditions for the development of cancer, or the standard of cancer care has been promoted over the years based primarily on pathophysiological considerations under only laboratory settings. Clinical trials involving supplementation of pharmacologic doses may reveal detrimental effects. Targeted therapies provide much more effective treatment for specific molecularly defined NSCLC subsets. The contradiction was due to the development of NSCLC studies. For example,

    Topic – 4 [cetuximab – TREATS/NEG_TREATS - Non-Small Cell Lung Carcinoma]

    *The use of **cetuximab**, a monoclonal antibody targeting the epidermal growth factor receptor (EGFR), has the potential to **increase survival** in patients with advanced **non-small-cell lung cancer**.* [PMID: 19410716]

    *Expert opinion: **Cetuximab** currently **has no role** in **NSCLC** treatment outside of research settings.* [PMID: 29534625]

*(4) Cost-effectiveness.* Cost-effectiveness is also an important concern. For cost-saving, the optimal strategy is to abandon ineffective medical practices. Lung cancer accounts for 20% of all cancer care budgets with limited benefits, besides the major source of morbidity and mortality, but also the healthcare costs which was associated with the diagnosis, staging, and management. For cost-effectiveness, it's necessary to identify patients in potential risks and tailoring the application of follow-up to the estimated risk individually.

*(5) Short-term outcome.* In clinical trials of cancer agents, the endpoint might be mortality, decreased pain, or the absence of disease. Five-year survival rate as surrogate endpoints may be used instead of stronger indicators to predict the clinical benefits, including a shrinking tumor or lower biomarker levels, because the results of the trial can be measured sooner. The following example has shown that Lung Diseases are treated as adverse events (surrogate endpoints) of erlotinib use.

    Topic – 11 [erlotinib – PREDISPOSES/NEG_AUGMENTS - Lung Diseases, Interstitial]

    *Our meta-analysis has demonstrated that **erlotinib**, gefitinib, and afatinib are associated with an **increased risk** of high-grade **interstitial lung disease** in patients with NSCLC.* [PMID: 25804125]

    *The addition of **erlotinib** to chemotherapy was well tolerated, with no increase in hematologic toxicity, and **no treatment-related interstitial lung disease**.* [PMID: 19738125]

*(6) Publication bias, citation bias, and time-lag bias.* Publication bias occurs when positive trials involving a medical intervention have been publicized more than neutral or negative trials of similar quality. Since specialist articles apparently continued to cite the studies that supported their own lines of research, the presence of refuting data was not mentioned in many articles.

*(7) Inaccuracy of SemRep.* Our analysis was based on semantic triples captured by the SemRep from the mining of biomedical literature abstracts. Some groups of contradictions were due to semantic differential was not correct. The parser identified the phrases that are mapped to UMLS Metathesaurus concepts by the MetaMap program (Aronson and Lang, 2010). Some studies evaluated the precision and recall of the predication of SemRep, the SemRep's precision for clinically relevant predictions is estimated at about 75%, and the recalls were reported around 50%. The following example has shown the SemRep precision error.

    Topic - 3 [Carboplatin - TREATS - Thrombocytopenia]

    *Conversely, Grade 3-4 **thrombocytopenia** was **more common** ($P < 0.0009$) and platelet transfusion was more frequent ($P < 0.05$) with **carboplatin** therapy.* [PMID: 11505405]

In addition, after eliminated inaccurate semantic predications generated by SemRep, the entire pair of contradictory knowledge were filtered out. And these drug-disease knowledge claims were further identified and 6 of them were classified into the class of diversity.

### 4.4.3 Identified drug-disease knowledge with diversity

54 pairs of knowledge were manually identified from 66 candidates automatically discovered by proposed algorithm. Therefore, the number of knowledge pairs with diversity will be 60 (e.g., 54 + 6), while 6 were identified through manual biocuration of automatic discovery of knowledge with contradiction. For the drug-disease knowledge with diversity, we investigated their diverse predicates (Table 8). Clearly, the three predominated groups are PREVENTS&TREATS, CAUSES&PREDISPOSES and DISRUPTS&TREATS.

Table 8. Identified diverse knowledge with their diversity labels

|   | **Diversity** | # |
|---|---|---|
| 1 | PREVENTS, TREATS | 23 |
| 2 | CAUSES, PREDISPOSES | 15 |
| 3 | DISRUPTS, TREATS | 11 |
| 4 | AUGMENTS, CAUSES, PREDISPOSES | 3 |
| 5 | AUGMENTS, PREDISPOSES | 3 |
| 6 | DISRUPTS, PREVENTS | 2 |
| 7 | AUGMENTS, NEG_PREVENTS, PREDISPOSES | 1 |
| 8 | NEG_PREDISPOSES, PREVENTS | 1 |
| 9 | DISRUPTS, PREVENTS, TREATS | 1 |

## 5. Discussion

### 5.1 Measuring medical knowledge based on SPO triples

In this paper, we have introduced a conceptual framework for measuring medical knowledge (or called "medical knowmetrics"), which takes SPO triples as the knowledge unit and the uncertain information as the context. The textual uncertainty associated with such SPO triples has been detected from abstract sentences in scientific literature.

The knowledge unit is a crucial component of knowledge. It is the smallest unit that can be comprehended by computer to define knowledge in a particular field. (Quigley & Debons, 1999) offers an interrogative-based approach to differentiating and quantifying information and knowledge within text. Knowledge is text that answers how/why in the problem space; Information is text that answers when/where/who/what in the problem space; Data is text that answers no question in the problem space. (Sidi, Jabar, Selamat, Ghani, & Sulaiman, 2009) then presents an Interrogative Knowledge Identification framework to identify unstructured documents that encompassed knowledge, information, and data, to identify knowledge from unstructured document in order to extract them. Briefly restating the interrogative knowledge identification, it identifies the type of document by separation of text into knowledge, information or data and unifying it with personal components of values and beliefs. The approach of answering interrogatively is used to answer the question within the text in unstructured document to identify knowledge. (Ding et al., 2013) proposed the new concept of knowledge entities, which is as carriers of knowledge units in scientific articles and include such entities as, keywords, topics, subject categories, datasets, key methods, key theories, and domain entities (e.g., biological entities: genes, drugs, and diseases). While the concept of entitymetrics has extended research objects from scientific literature to knowledge entities, the entity co-occurrence network and entity citation network only show a correlation between entities, not involving their causal nature. For example, for the co-occurrence between drugs and diseases, we do not know whether it is referring to the "therapeutic use" or "induced adverse event" of a given drug for a given disease. We argue that the semantic predications based on the "entity-relationship-entity" triples answer the questions of "how", and such triples associated to a given particular topic which formed the knowledge graph answer the questions of "why" in the framework by (Quigley & Debons, 1999). For example, consider this sequence of two relations (obtained from SemMedDB) with "stimulates" and "treats" predicates in the following order: Mercaptopurine→*stimulates*→Cytarabine triphosphate→*treats*→ Leukemia. This subgraph explains why mercaptopurine can treat Leukemia.

From the above discussion, it is reasonable for measuring medical knowledge at the level of semantic predications, i.e., Subject-Predicate-Object (SPO) triples, which can be generated by SemMedDB, a PubMed-scale repository of biomedical semantic predications.

## 5.2 Measuring the SPO triples related textual uncertainty

Currently, there are only a few studies on measuring the uncertainty level of scientific text surrounding the resulted SPO triples. In one chapter of their book, (C. Chen & Song, 2017a) analyzed conflicting and contradictory sentences and their SPO triples in SemMedDB, which states that these sentences have more "active" semantic relationships including "treat", "influence", and "prevent". The SemRep project team proposed a method to assign factuality values for the semantic predications extracted from medical literature to reflect the uncertainty level of knowledge, i.e., PROBABLE, POSSIBLE, DOUBTFUL, COUNTERFACT, UNCOMMITTED, and CONDITIONAL (H. Kilicoglu, Rosemblat, & Rindflesch, 2017).

There are also a few studies on identifying contradictory medical knowledge from cross-sentences based on semantic predication rules that the subject and object are the same but the predicate is just opposite between two or more SPO triples. For example, (Alamri, 2016) classified the predicates into three types: (a) active/causing, such as AUGMENTS and CAUSES; (b) passive/suppressing, such as DISRUPTS and PREVENTS; and (c) others, such as ADMINISTERED_TO and OCCURS_IN. If any of these types of relation pairs or their negative relationship can be extracted from two or more sentences, such as CAUSES and NEG-CAUSES, or active/causing and passive/suppressing are extracted, respectively, such as CAUSES and PREVENTS, the knowledge coming from the involved sentences is contradictory. Similarly, the SemRep project team adopted this rule to identify contradictory medical knowledge in clinical research papers. They identified clinically relevant predicate pairs and focused on (a) predicates with causal meaning, such as TREAT versus CAUSES, PREVENTS versus CAUSES, TREATS versus PREDISPOSES, and PREVENT versus PREDISPOSES; and (b) predicates without causal meaning, including TREATS, PREVENTS, CAUSES, PREDISPOSES and their negative-form predicates (Rosemblat et al., 2019). In order to identify controversial claims which semantically contradict each other, (Pinto, Wawrzinek, & Balke, 2019) focus on the following associations that are part of the semantic relations in UMLS: "affects", "associated-with", "causes", "inhibits", "prevents", "process-of", "treats" as well as its corresponding negative counterparts, e.g., "neg-affects", "neg-associated-with". However, they only scratched the surface of "controversy" using direct negations of the semantic orientation.

Concerning the above research, the uncertainty framework is somewhat complicated, and it is sometimes difficult to distinguish during manual annotations, such as "probably" and "possible". The classification of uncertainty needs to be simplified. Contradictory knowledge is identified based on the rule "with the same subject and object but opposite predicate" in semantic predications, but this rule ignores the textual uncertainty of the source sentence. To fill out the above gap, in our study, we simplified the classification of the uncertainty of medical knowledge, by focusing on only two types, i.e., 'diversity' and 'controversy/contradiction'. And we have filtered out the semantic predications interpreted from uncertain claims by using 'hedging' before detecting contradictory knowledge from cross-sentences. In addition, the scientific article in the health sciences evolved from the letter form and purely descriptive style in the seventeenth century to a very standardized structure in the twentieth century known as introduction, methods, results, and discussion (IMRaD) (Sollaci & Pereira, 2004). Semantic predications extracted from all the IMRaD sentences of a given paper may cause redundant. For example, the identical SPO triples from sentences in the title, the objective and the conclusion in the abstract tend to be extracted, whereas the knowledge assertion of a research is most important in the conclusive sentences. The SPO triples derived only from the conclusive sentences in the abstract may be distinguish for knowledge discovery in our future analysis.

## 5.3 Potential applications of detecting the uncertainty of medical knowledge

Firstly, measuring knowledge with SPO triples as the units and uncertainty as context could promote analytics from correlation to causality, and suggests innovative schemes for scientific frontier recognition. The SPO triples are the indivisible minimum knowledge units that define the causal relationship between entities, such as the therapeutic relationship (e.g., therapeutic effect) or cause (e.g., side effect) between drugs and diseases. The correlation is not concept co-occurrence, but is the core to extend from bibliometrics to entitymetrics and further deepen to knowledge metrics. The frontier of science generally comprises a high degree of uncertainty and competitive knowledge claims: unsolved scientific problems, and unconfirmed controversial research outcomes. Digging the knowledge that is at the stage of speculation, even contradiction or disputation provide a novel method to identify the frontier of science, which is different from the common practice in current information science research by using the highly cited papers.

Secondly, introducing the perspective of "solving knowledge gap and reducing knowledge uncertainty" may provide new solutions for research evaluation. By proposing indicators that measure the uncertainty level of knowledge units, one can reveal the extent to which a given scientific research diminishes the uncertainty for scientific problems in this field, such as verifying hypotheses or speculations, and resolving contradictions or disputes. This perspective highlights the value of the produced knowledge in solving scientific problems, and it will help reform the prevailing research evaluation mechanism that too much focusing on publications, citations and journals' impact factors.

Lastly, a measure to quantify the uncertainty of a given SPO triple could be developed by counting the frequency of uncertain sentences across all sentences. We can estimate the probability of the certainty level for a given knowledge claim. Medical science is a science of uncertainty, and its knowledge is often ambiguous, inconsistent, or even inaccurate. Medical decision-making can only be based on the limited, uncertain knowledge in reality. For specific medical problems or decision-making requirements (e.g., which drugs can be effectively used to treat lung cancer), if we can distinguish the related knowledge claims with higher certainty level from those with lower level, it will help to improve the efficiency of computational knowledge-driven decision support.

## 6. Conclusion

We have proposed a conceptual framework of *Medical Knowmetrics* by using semantic predications as the knowledge unit and the uncertainty as the knowledge context. And a case study on extracting and characterizing clinical knowledge with 'diversity' and 'controversy or contradiction' on drug treatment for lung cancer from large-scale published medical documents has validated the proposal. The uncertainty of scientific knowledge and how its status evolves over time indirectly reflect the strength of competing knowledge claims, the contribution for filling up knowledge gap, as well as the probability of certainty for a given knowledge claim. So, we try to provide new insights using the uncertainty-centric approaches to detecting research fronts, evaluating academic contributions and improving the efficacy of computable knowledge driven decision support. We expect to deepen the methodologies from scientometrics and informetrics, to knowmetrics and broaden their new application fields. In the future, we will extend the source sentences to citing sentences. In our opinion, from the perspective of textual analysis of uncertainty in citing sentences, we can capture how scholars evaluate the reliability and validity of relevant knowledge claims. This could explain the evolutionary process of scientific knowledge more effectively.

## Acknowledgments

Xiaoying Li and Suyuan Peng contributed equally. Jian Du proposed the conceptual framework, designed the research and wrote the manuscript. Xiaoying Li conducted the data analysis and Suyuan Peng interpreted and characterized the results. Both authors participated in writing the manuscript. This work was funded by the National Natural Science Foundation of China (71603280), and the Young Elite Scientists Sponsorship Program by China Association for Science and Technology (2017QNRC001).